\documentclass[10pt,twocolumn,letterpaper]{article}

\usepackage{wacv}              

\usepackage{graphicx}
\usepackage{amsmath}
\usepackage{amssymb}
\usepackage{booktabs}

\usepackage{algorithm}
\usepackage{algpseudocode}

\usepackage{adjustbox}
\usepackage{multirow}
\newcommand{\boldhline}{\specialrule{0.15em}{0em}{0.1em}}

\makeatletter
\DeclareRobustCommand\onedot{\futurelet\@let@token\@onedot}
\def\@onedot{\ifx\@let@token.\else.\null\fi\xspace}

\def\eg{\emph{e.g}\onedot} 
\def\ie{\emph{i.e}\onedot} \def\Ie{\emph{I.e}\onedot}
\def\cf{\emph{cf}\onedot}

\def\etal{\emph{et al}\onedot}
\makeatother

\usepackage[pagebackref,breaklinks,colorlinks]{hyperref}

\usepackage[capitalize,sort&compress,nameinlink]{cleveref}
\crefname{section}{Sec.}{Secs.}
\Crefname{section}{Section}{Sections}
\Crefname{table}{Table}{Tables}
\crefname{table}{Tab.}{Tabs.}

\usepackage[accsupp]{axessibility}

\def\wacvPaperID{1265} 
\def\confName{WACV}
\def\confYear{2025}

\begin{document}

\title{Modality-Incremental Learning with Disjoint Relevance Mapping Networks for Image-based Semantic Segmentation}

\author{
	Niharika Hegde $^{1,2\hspace{0.25em}*}$ \qquad Shishir Muralidhara $^{2\hspace{0.25em}*}$ \qquad René Schuster $^{1,2}$ \qquad Didier Stricker $^{1,2}$\\
	$^1$ RPTU -- University of Kaiserslautern-Landau \\
	$^2$ DFKI -- German Research Center for Artificial Intelligence \\
	{\tt\small firstname.lastname@dfki.de}
}

\maketitle

\def\thefootnote{*}\footnotetext{\hspace{0.25em}These authors have contributed equally to this work.}

\begin{abstract}
   In autonomous driving, environment perception has significantly advanced with the utilization of deep learning techniques for diverse sensors such as cameras, depth sensors, or infrared sensors. 
   The diversity in the sensor stack increases the safety and contributes to robustness against adverse weather and lighting conditions.
   However, the variance in data acquired from different sensors poses challenges.
   In the context of continual learning (CL), incremental learning is especially challenging for considerably large domain shifts, e.g. different sensor modalities.
   This amplifies the problem of catastrophic forgetting.
   To address this issue, we formulate the concept of \emph{modality-incremental learning} and examine its necessity, by contrasting it with existing incremental learning paradigms.  
   We propose the use of a modified Relevance Mapping Network (RMN) to incrementally learn new modalities while preserving performance on previously learned modalities, in which relevance maps are disjoint.
   Experimental results demonstrate that the prevention of shared connections in this approach helps alleviate the problem of forgetting within the constraints of a strict continual learning framework.
\end{abstract}

\section{Introduction} \label{sec:intro}
Continual learning (CL) has emerged as a fundamental paradigm to address the need for intelligent agents to continually update with new information while preserving learned knowledge. 
In contrast, conventional machine learning normally builds on a closed dataset, \ie it can only handle a fixed number of predefined classes or domains, and all the data needs to be presented to the model in a single training step.
However, in practical scenarios, models frequently face the challenge of dealing with changing data and objectives.
This problem can be circumvented by accumulating all data and retraining the model to derive a unified model effective across a combined dataset.
Although this approach achieves optimal performance, it is often impractical and may not be feasible due to several reasons.
For instance, anticipating future data is not possible in real-world applications, and access to previous data might be restricted due to privacy concerns or resource constraints. 
Moreover, retraining from scratch using all past data results in a significant increase in training time and computational requirements.
Consequently, learning solely from new data is more efficient, but can lead to catastrophic forgetting \cite{catastrophic_forgetting}, where past knowledge is overwritten resulting in degraded performance on the previous tasks. 
This challenge emphasizes the importance of developing CL methods to maintain a balance between incorporating new information and retaining past knowledge, referred to as the stability-plasticity dilemma \cite{stability_plasticity}.

Autonomous driving systems are typically trained on normal driving conditions due to their prevalence and ease of accessibility. 
However, as these systems advance, they must confront a multitude of driving scenarios, including adverse weather, low-light conditions, and other challenging environments. 
This shift in data distribution, can undermine their ability to make precise predictions or decisions, raising potential safety concerns.
Single sensor systems, in particular, struggle to adapt to challenging conditions which can severely impact their performance.
Integrating a multi-modal, complementary sensor suite is an effective measure to encounter deficiencies under such changes of conditions.
For example, IR cameras are effective under low-light conditions but can be affected by weather conditions like rain and fog.
Depth sensors offer precise distance measurements but may be limited in range. 
Combining diverse sensors in a heterogeneous stack helps alleviate the limitations of individual sensor types and enhances the overall performance and reliability of autonomous systems.

For an existing system, new sensor modalities might be introduced as they undergo technical advancements, become more cost efficient, or address specific limitations.
In such cases, it's appealing to have a single, unified model that incrementally learns to handle the new modalities and enhances its ability to perceive under challenging driving conditions and varying sensor characteristics, without forgetting previously acquired knowledge.
In this paper, we introduce and formalize this novel incremental setting termed \textit{modality-incremental learning} (MIL) to learn on an extending set of sensor modalities and contrast it against existing incremental paradigms. 
We exemplify the concept of MIL by semantic segmentation on various visual modalities (\ie RGB, IR, and depth cameras) in an automotive setting.

Current incremental settings typically use data from a single visual modality, and the methods designed for them lack the capability to manage changing modalities. 
Addressing this challenge of learning visual modalities, we propose the use of Disjoint Relevance Mapping Networks (DRMNs), which aim to learn an improved representational map, such that the significantly distinct tasks (changing modalities) use different subsets of the network parameters.
We argue that the prevention of overlap in the relevance maps mitigates forgetting completely, without having a negative impact on the utilized network's capacity.
The contribution of our work can be summarized as follows:
\begin{itemize}
    \item We introduce and formulate the problem of modality-incremental learning (MIL) in the context of continual learning, and demonstrate it for semantic segmentation in an automotive context.
    \item We benchmark existing methods for domain-incremental learning (DIL) in this novel setting.
    \item We propose a modified version of Relevance Mapping Networks (RMN) \cite{RMN} that is tailored towards MIL.
    \item We evaluate the proposed Disjoint Relevance Mapping Networks (DRMN) in terms of accuracy, forgetting, and network utilization on various MIL settings across two multi-modal datasets.
\end{itemize}

\section{Related Work} \label{sec:related}
Continual learning strategies can be categorized into three types: Architecture-based, replay, and regularization methods.
Architecture-based methods address forgetting by altering the architecture of networks either explicitly or implicitly to learn new tasks.
Explicit modification involves dynamically expanding the network architecture by adding individual neurons \cite{DEN}, widening/deepening layers \cite{growing_brain}, or cloning the network \cite{progressive_neural_networks}.
Implicit modifications use a fixed network capacity and adapt to new tasks through freezing \cite{network_freezing}, pruning \cite{packnet} or task-specific paths \cite{pathnet}.
Architecture-based methods also include dual-architecture models inspired by the brain \cite{bio_arch,fearnet}.

Replay-based methods address forgetting by replaying previously encountered information. These methods can be classified into experience replay and generative replay. 
Experience replay \cite{experience_replay, rehearsal_rebalancing} or rehearsal, involves storing a subset of instances from the previous task, which are later used during retraining on a new task. 
However, experience replay faces challenges related to privacy and storage of data.
Generative replay \cite{DGR, memoryGAN} methods diverge from rehearsal approaches by training generative models, allowing them to generate samples from previous tasks. 

Regularization is a process of introducing an additional term into the loss function to regulate the update of weights when learning in order to retain previous knowledge. 
Regularization includes identifying crucial weights \cite{EWC,SI,MAS} within a model and preventing overwriting them, or storing learned patterns to guide the gradients \cite{conceptors, conceptor_patterns}.
Distillation methods \cite{born_again, ILT} transfers knowledge from one neural network to another. Such methods do not need to store data, and only require a previous model for knowledge transfer.

In this work, we propose a hybrid approach that builds on RMNs \cite{RMN} and combines architectural and regularization techniques. 
The idea is to maintain a fixed network capacity by freezing task-specific weights and utilize pruning to free weights for subsequent tasks. 
The relevance maps help in identifying the important weights from previous tasks, and we enforce parameter isolation by masking these weights.

\subsection{Continual Semantic Segmentation}
Continual semantic segmentation (CSS) constitutes a specialized sub-field within the broader realm of continual learning, focusing specifically on semantic segmentation.
Most research in CSS follows either one out of two popular incremental learning schemes.
The first is class-incremental learning (CIL) \cite{MIB,PLOP,AWT,RCIL,SSUL}, in which sets of classes are learned sequentially.
The second is domain-incremental learning (DIL), which is closer to the proposed MIL setting.
Here, the distribution of input data is extended over time.
In fact, MIL can be viewed as a severe form of DIL, in which individual sensor modalities represent entirely different visual domains.
For domain-incremental semantic segmentation, MDIL \cite{MDIL} partitions the encoder network into domain-agnostic and domain-specific components to learn new domain-specific information, and a dedicated decoder is instantiated for each domain.
DoSe \cite{Dose} uses domain-aware distillation on batch normalization for incremental learning using a pretrained model.
It also uses rehearsal for storing and replaying difficult instances from previously seen domains.
Addressing the storage constraints in rehearsal-based approaches, Deng and Xiang \cite{StyleReplay} propose a style replay method to reduce storage overhead.  

Our work is in contrast with the existing work by Barbato \etal \cite{mil_cil} who use multiple modalities in a continual learning setting within the context of CIL.
\Ie, all modalities are used in all tasks.
Their work assumes a pre-defined number of modalities, allowing for the design of suitable architectures.
MIL in this work aligns more closely with DIL since the number of classes remains consistent across tasks.

\subsection{Multi-Modal Semantic Segmentation}
Early multi-modal segmentation methods \cite{indoorseg_early_fusion} combined data from different modalities and used this combined input for the segmentation network. 
However, this strategy of early fusion struggles to effectively capture the diverse information provided by different modalities.
Recent advancements aim to leverage the strengths of various modalities by employing multiple fusion operations at various stages of the network \cite{fusenet}. 
A common architectural choice involves a multi-stream encoder \cite{rfbnet}, where each modality has its own network branch. 
Additional network modules \cite{acnet} connect these branches to combine modality-specific features across branches, facilitating hierarchical fusion. 

For multi-modal segmentation using RGB and depth modalities, AsymFusion \cite{AsymFusion} uses a bidirectional fusion scheme with shared-weight branches and asymmetric fusion blocks to
enhance feature interactions.
Chen \etal \cite{SAGate} proposed a unified cross-modality guided encoder with a separation-and-aggregation gate (SA-Gate) for effective feature re-calibration and aggregation across modalities
Mid-fusion architecture \cite{mid_fusion} combines sensor modalities at the feature level using skip connections for autonomous driving.
CMX \cite{CMX} leverages cross-modal feature rectification and fusion modules, integrating a cross-attention mechanism for enhanced feature fusion across modalities.

For multi-modal segmentation using RGB and IR modalities, ABMDRNet \cite{abmdrnet}, uses a bi-directional image-to-image translation to mitigate modality differences between RGB and thermal features.
GMNet \cite{GMNet} integrates multi-layer features using densely connected structures and residual modules, with a multistream decoder that decouples semantic prediction into foreground, background, and boundary maps.
RTFNet \cite{RTFNet} characterized by the asymmetrical encoder and decoder modules, merges modalities at multiple levels of the RGB branch. 
FuseSeg \cite{FuseSeg} proposed the hierarchical addition of thermal feature maps to RGB feature maps in a two-stage fusion process. 
CCAFFMNet \cite{CCAFFMNet} leverages multi-level channel-coordinate attention feature-fusion blocks within a coarse-to-fine U-Net architecture.

This work addresses multi-modal segmentation from a continual learning perspective, where modalities are incrementally and arbitrarily added.
This complicates the design of specialized architectures for handling multiple modalities.
Therefore, we process each modality independently for segmentation, leaving more advanced fusion techniques to the possibilities for future research.

\begin{figure}
    \centering
    \includegraphics[width=0.32\columnwidth]{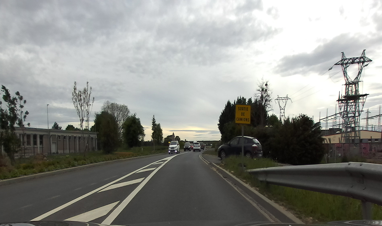}
    \includegraphics[width=0.32\columnwidth]{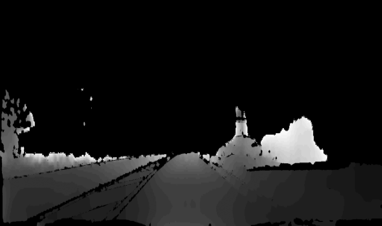}
    \includegraphics[width=0.32\columnwidth]{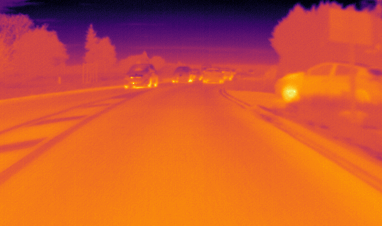}
    \caption{Three different modalities to perceive traffic scenarios in an automotive context. From left to right: Classical RGB, depth, and IR images from the InfraParis dataset \cite{infraParis}.}
    \label{fig:modalities}
\end{figure}

\section{Modality-Incremental Learning (MIL)} \label{sec:mil}

Incremental learning involves learning a sequence of tasks $T = T_0, T_1,...,T_n$. 
Each task $T_i$ is associated with task-specific data $D_i = (X_i, Y_i)$, and represents a change either in the input or the output distribution.
In domain-incremental learning (DIL), the input distribution $X$ changes at each task increment, while the output distribution remains the same. 
Each task can represent different data sources such as geographical locations or weather conditions.
In class-incremental learning (CIL), the input data remains constant, while each task introduces a subset of new classes $C_i$, such that $C_0 \cup C_1 \cup C_i = C \in Y$ the model has to learn without forgetting previously learned classes.

We introduce modality-incremental learning (MIL), a novel incremental learning setting tailored to handle the case of incrementally learned sensor modalities.
In MIL, each new task with associated data $(M_i, Y)$ presents a change in the input distribution by introducing a new modality $M_i$.
The set of classes $Y$ remains consistent across all tasks, similar to DIL.
Unlike DIL, where the input $X_i$ remains within the same visual modality across all tasks, MIL faces more significant data drift due to the introduction of new modalities.
As a result, DIL methods struggle to adapt effectively to MIL scenarios, as shown in our experiments.

To underline this difference and the severity of the domain gaps between modalities, we highlight that even an offline training on joint data from all MIL tasks produces subpar results compared to modality-specific modules.
In CL, this \textit{joint training} usually forms a theoretical upper bound, since diverse data facilitates the learning and forgetting does not occur.
However in MIL, the substantial differences between modalities pose a significant challenge for joint training to effectively utilize shared knowledge across tasks.

A notable advantage of MIL, compared to DIL or CIL, is the straightforward availability of the task ID, as it can be safely assumed that the sensor that produces the input signal is known to the system. 
This inherent knowledge obviates the need for explicit task identification during inference.

\begin{figure*}
    \centering
    \begin{subfigure}[b]{0.45\textwidth}
        \centering
        \includegraphics[width=\linewidth]{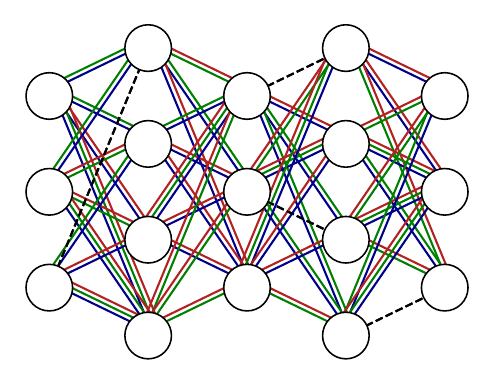}
        \caption{Relevance Mapping Network (RMN)}
        \label{fig:rmn_architecture}
    \end{subfigure}
    \begin{subfigure}[b]{0.45\textwidth}
        \centering
        \includegraphics[width=\linewidth]{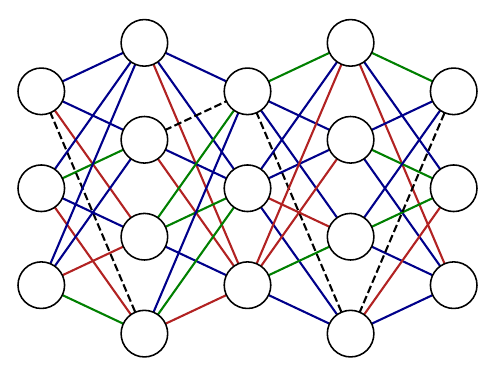}
        \caption{Disjoint Relevance Mapping Network (DRMN)}
        \label{fig:drmn_architecture}
    \end{subfigure}

    \vspace{1em} 
    
    \begin{subfigure}[b]{0.65\textwidth}
        \centering
        \includegraphics[width=\linewidth]{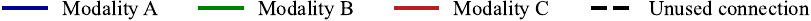}
    \end{subfigure}
    
    \caption{Relevance Mapping Network (RMN) (left) shares connections across tasks, with new tasks utilizing their respective relevance map values and the previous weights. In contrast, the Disjoint RMN (DRMN) (right) isolates connections between tasks, compelling the network to learn independent, task-specific weights and mitigates the negative interference when incrementally learning modalities. It is important to note that each node can be used for a modality-specific representation in all tasks.}
    \label{fig:rmn_drmn_architecture}
\end{figure*}

\section{Disjoint Relevance Mapping Networks} \label{sec:drmn}
The challenge of learning multiple modalities lies in the inability of a single encoder to manage them all, even in an offline setting, and this is exacerbated when learning modalities incrementally.
This limitation renders most continual learning methods, such as distillation and rehearsal or replay, ineffective as they still rely on a single network. 
Architecture-based methods, such as multi-encoder or multiple networks show promise but do not scale well with an increasing number of tasks.
The number of models and storage requirements grow proportionally with each new modality.
In light of these limitations, it would be desirable to have a single model of fixed size that can effectively handle various modalities, unlike previous methods. 
To this end, we propose using Relevance Mapping Networks (RMNs) \cite{RMN} to handle incremental learning of modalities.
This approach requires the task ID to be known during inference, which is not an issue in MIL as explained earlier.
We further modify the original RMN concept with parameter isolation to better fit the needs for MIL. 

\subsection{Relevance Mapping Networks} \label{sec:drmn:rmn}
\begin{figure}[t]
	\centering
	\includegraphics[width=0.75\columnwidth]{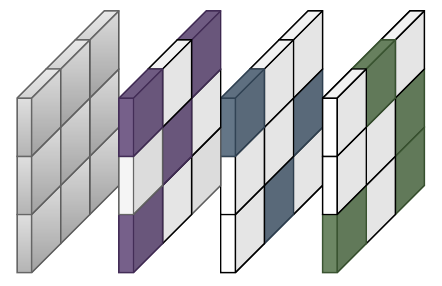}
	\caption{A Relevance Mapping Network augments the network weights by adding task-specific relevance maps $\mathbb{M}$ to select the important weights for each task.}
\label{fig:adjx_maps}
\end{figure}

RMNs are a method inspired by the optimal overlap hypothesis, which aim to learn an optimal representational overlap, such that unrelated tasks use different network parameters, while allowing similar tasks to have a representational overlap.
RMN was originally proposed for image classification in the continual learning setting \cite{RMN}. 
In this work, we extend the implementation of RMN beyond image classification, to tackle the complex task of continual semantic segmentation. 
RMNs enhance existing neural networks by augmenting the convolutional and linear layers with additional weights referred to as relevance maps $\mathbb{M}$ as illustrated in \cref{fig:adjx_maps}.
These relevance maps are unique to each task and identify the most important neural connections within the network for the corresponding task, and are used in conjunction with the standard layer weights $\textbf{W}$:

\begin{equation}
    f_{out} = \textbf{W} \cdot \mathbb{M}_{t} \cdot f_{in}
\end{equation}

The learned relevance maps can be interpreted as (soft) masks, selecting and freezing the crucial task-specific weights in the network, resulting in dynamic task-specific paths. 
This approach effectively dissects the network into partial subnetworks, while still allowing it to share information across related tasks and maintaining task-specific weights. 

The relevance maps $\mathbb{M}$ are learned continuously with a bounded activation in the interval $[0..1]$.
During training, the RMN periodically applies thresholding with a hyperparameter $\mu$, called the pruning parameter, to select relevant connections, of which the corresponding values in $\textbf{W}$ are frozen, and to set irrelevant connections to zero (both in $\mathbb{M}_t$ and $\textbf{W}$). 
This way, unused capacity of the network is freed for future tasks.

\subsection{Disjoint Relevance Mapping Networks for MIL} \label{sec:drmn:method}

A key challenge in using RMNs for modality-incremental learning is to balance the utilization of network capacity and the overlap between tasks.
Especially in MIL, we argue that too much overlap between the task-specific network paths hinders learning and amplifies forgetting, due to the naturally large differences between modalities (\cf \cref{sec:mil}).
With regular RMNs, we observe a significant overlap of used connections across tasks.
While the freezing of relevant weights helps retain knowledge from previous tasks, it lacks a mechanism to promote increased adaptation to drastically different sensing modalities.
Too much overlap in relevance maps forces the networks to reuse the previously learned weights, and with highly disparate modalities, this leads to inadequate learning on the new task.
We demonstrate under \cref{sec:experiments:shared_weights}, the paradoxical effect of weight sharing, which typically is beneficial for knowledge transfer, but becomes detrimental in this context due to modality-specific conflicts. 

To address this issue of overlap, we propose Disjoint Relevance Mapping Networks (DRMNs). 
As the name suggests, DRMNs enforce a complete separation of relevant neural connections between modalities.
The idea and differences to classical RMNs are visualized in \cref{fig:rmn_drmn_architecture}.
RMN allows sharing connections across tasks, with new tasks potentially reusing the previously learned connections with the newly learned relevance maps. 
In contrast, DRMN uses parameter isolation, and the network is forced to learn task-specific connections for each new task.
By doing so, DRMN aims to reduce interference and conflicts that can arise when learning diverse modalities.
To enforce parameter isolation, each relevance map of the previous tasks $\mathbb{M}_i \forall i<t$ is analyzed to identify used connections.
These values are then set to zero in $\mathbb{M}_t$, effectively rendering them unimportant for the current task.
This limits the learning of $\textbf{W}$ and $\mathbb{M}_t$ to connections not used in prior tasks.
Intuitively, one might assume that this strict separation has a negative impact on the transfer of knowledge and the depletion of the network's capacity.
However, our experiments in \cref{sec:experiments:util,sec:experiments:benchmark} show that this is not the case.
While connections must not be shared across tasks, network nodes can be used in multiple modalities. 
This mechanism effectively minimizes negative interference between tasks during learning, but allows to learn powerful representations from sparse connections.
In fact, forgetting is reduced due to the increased decoupling, while the overall utilization of connections is barely affected, compared to original RMNs.
The incremental training process with DRMNs is detailed in \cref{alg:drmn}.

\begin{algorithm}[t]
	\caption{Disjoint Relevance Mapping Network}
	\begin{algorithmic}[1]
		\State \textbf{Training Phase}
		\State \textbf{Input}: Training data $(X, Y)$, task IDs $t = 0,1, \dots, n$, prune parameter $\mu$, initialized relevance maps $\mathbb{M}$
		\For{$t=0$ to $n$} \Comment Train on task $t$
                \If{$t>0$}
                    \State $m_\textbf{unused} \gets \bigwedge_{i=0}^{t-1}(\neg \mathbb{M}_i)$
                \Else
                    \State $m_\textbf{unused} \gets \mathbf{1}$
                \EndIf
        		\State $f(X_t; \textbf{W}; \mathbb{M}_{t}) \implies \hat{Y}_t \gets \sigma((\textbf{W} \cdot \mathbb{M}_{t} \cdot m_\textbf{unused}) \cdot X_t)$
        		\State Relevance Map Pruning, $\mathbb{M}_t \leq \mu$
    		\State Freeze weights in $f$ where $\mathbb{M}_t \neq 0$
		\EndFor
		\State \textbf{Inference Phase}
		\State \textbf{Input}: Task ID $t$ is given by the sensor and used to select the relevance map $\mathbb{M}_t$ for that task. 
		\State \textbf{Output}: $f(X; \textbf{W}, \mathbb{M}_t)$
	\end{algorithmic}
	\label{alg:drmn}	
\end{algorithm}

\section{Experiments} \label{sec:experiments}

\begin{table*}[t]
	\caption{Results for three MIL task sequences on Freiburg Thermal \cite{freiburgThermal} dataset after learning all tasks.}    
	\begin{adjustbox}{width=\textwidth}
		\begin{tabular}{c||c|c|c|c||c|c|c|c||c|c|c|c}
			\boldhline
			\multirow{2}{*}{\textbf{Method}} & \multicolumn{4}{c||}{\textbf{RGB→IR→Gray}}  & \multicolumn{4}{c||}{\textbf{Gray→RGB→IR}} & \multicolumn{4}{c}{\textbf{IR→Gray→RGB}} \\ \cline{2-13}
			& RGB & IR & Gray & \textbf{Avg} & Gray & RGB & IR & \textbf{Avg} & IR & Gray & RGB & \textbf{Avg} \\ \boldhline
			\textbf{Single Task}   & 76.41  & 59.56  & 74.56  & \textit{70.18}  & 74.56  & 76.41  & 59.56  & \textit{70.18}& 59.56  & 74.56  & 76.41  & \textit{70.18}      \\ \hline
			\textbf{Joint Training}    & 75.43  & 56.06  & 74.46  & \textit{68.65} & 74.46 & 75.43  & 56.06  & \textit{68.65}& 56.06  & 74.46  & 75.43  & \textit{68.65}       \\ \boldhline
			\textbf{Fine Tuning}       & 74.97  & 07.41  & 74.88  & \textit{52.42}  & 13.76  & 14.06  & 60.19  & \textit{29.34} & 07.12  & 73.17  & 75.24 &  \textit{51.84}       \\\hline
			\textbf{EWC} \cite{EWC}    & 72.28  & 07.98  & 72.93  & \textit{51.06}  & 18.38  & 18.02  & 40.80  & \textit{25.73} & 10.56  & 59.47  & 61.69 &  \textit{43.91}       \\ \hline
			\textbf{ILT} \cite{ILT}    & 74.02 & 07.91  & 66.27  & \textit{49.40}  & 13.44  & 13.40  & 08.96  & \textit{11.93} & 20.68  & 22.57  & 23.53 &  \textit{22.26}    \\ \hline
			\textbf{RMN} \cite{RMN}    & 73.13  & 55.01  & 68.29  & \textit{65.48}  & 71.09  & 72.82  & 53.57  & \textit{65.83} & 55.10  & 68.90  & 69.46 &  \textit{64.49}    \\ \hline
			\textbf{DRMN} (Ours)       & 73.21  & 54.95  & 69.38  & \textit{\textbf{65.85}}  & 71.12  & 72.61  & 54.12  & \textit{\textbf{65.95}} & 54.97  & 70.56  & 71.19 &  \textit{\textbf{65.57}}    \\ \boldhline
		\end{tabular}
	\end{adjustbox}
	\label{tab:freiburg_thermal}	
\end{table*}

\subsection{Datasets} \label{sec:datasets}
To simulate a sequentially adapted sensor system, we utilize datasets that offer ground truth for our task, \ie pixel-wise semantic labels, as well as input images captured with different sensors.
The datasets Freiburg Thermal \cite{freiburgThermal} and InfraParis \cite{infraParis} offer aligned RGB and infrared images.
The InfraParis dataset offers an additional visual modality in the form of depth maps, enhancing the diversity of available modalities. 
To further enhance the versatility of the experimental setup, an additional visual modality in the form of grayscale images was created for both the datasets. 
This diversity allows for a more comprehensive assessment of the proposed approaches for MIL, as well as facilitating the exploration of the effects of incrementally learned modalities. 

\begin{itemize}
	\item \textbf{Freiburg Thermal} \cite{freiburgThermal} encompasses diverse driving scenes such as highways, cities, suburbs, and rural areas with pixel-level labels for 13 object categories. Including the simulated grayscale sensor, it covers 3 modalities.
    The dataset has 9,735 images for training and 2,435 for validation for each modality.
	
	\item \textbf{InfraParis} \cite{infraParis} dataset contains RGB, depth, and infrared data captured in various cities around Paris. This yields 4 different sensor modalities, when augmented with grayscale images. The dataset offers pixel-wise annotations for 20 classes.	
    It contains 3545 RGB images as well as 6567 Depth and IR images (each) for training and 189 images per modality for validation.
    For depth images, we mask ground truth labels wherever the corresponding depth value is zero.
\end{itemize}

\subsection{Implementation and Baselines} \label{sec:experiments:baseline}
We evaluate the results of our approach against existing continual learning approaches and the standard baselines, \ie \textit{joint training}, \textit{fine-tuning}, and \textit{single-task learning}.  
Joint training learns the set of tasks concurrently in a single step, using all modalities.
Since the model is trained on all tasks simultaneously, there is no catastrophic forgetting and the learning benefits from the extended dataset size.
The single-task baseline refers to a model trained on just one task without any further adaptation or incremental steps.
In the context of MIL, the single-task models have been trained on individual modalities and serve as the upper bound for comparison.
Fine-tuning is a naive solution for incremental learning, in which a model is trained sequentially for each task, building on the previously learned tasks.
While effective for learning new tasks, it suffers most from forgetting of previous tasks, especially with modalities that are considerably different, \eg RGB and IR images.
In our experiments, we further compare against two regularization-based approaches, Elastic Weight Consolidation (EWC) \cite{EWC} which penalizes overwriting of important weights, and Incremental Learning Techniques (ILT) \cite{ILT} which uses knowledge distillation.
We use the Relevance Mapping Network (RMN) from \cite{RMN} adapted for segmentation as our baseline for comparison. 
The experiments highlight its shortcomings in handling multiple modalities and learning new modalities. 
Our proposed Disjoint RMN (DRMN) overcomes this limitation by ensuring each task learns independent weights.

For the segmentation network, we use a DeeplabV3+ \cite{deeplabv3plus} model with a ResNet-101 \cite{resnet} backbone, which is pretrained on the ImageNet \cite{ImageNet} dataset.
This backbone is used for all initial, single and joint-training models. 
In incremental learning methods, the previous task model serves as the starting point.
As part of our work, we use a custom DeepLabV3+ model that incorporates relevance maps into the convolutional layers, along with task-specific batch normalization \cite{RMN}.
All methods used for comparison are trained on a single RTXA6000 GPU with a batch size of 2 for 75 epochs, using the stochastic gradient descent (SGD) optimizer with a learning rate of 1e-5 across all tasks.
The prune parameter $\mu$ is used as a threshold for determining the important weights in the relevance maps.
We use $\mu=0.6$ and weights below this value are pruned after each epoch starting from epoch 50.
Results for different threshold values are provided in our supplementary material.
Our experiments are evaluated in terms of mean Intersection-over-Union (mIoU).

\subsection{Benchmark} \label{sec:experiments:benchmark}

\subsubsection{Freiburg Thermal} \label{sec:experiments:benchmark:freiburg}

Using the three modalities from the Freiburg Thermal dataset \cite{freiburgThermal}, we design the following non-exhaustive task sequences. 
These sequences cover all possible learning orders for each modality: (\textit{\textbf{RGB}} $\rightarrow$ \textit{\textbf{IR}} $\rightarrow$ \textit{\textbf{Gray}}), ~(\textit{\textbf{Gray}} $\rightarrow$ \textit{\textbf{RGB}} $\rightarrow$ \textit{\textbf{IR}}) and ~(\textit{\textbf{IR}} $\rightarrow$ \textit{\textbf{Gray}} $\rightarrow$ \textit{\textbf{RGB}}).

From \cref{tab:freiburg_thermal}, we can observe that the best results are achieved using the single-task models. 
The joint training baseline falls short due to the higher complexity of learning different modalities. 
However, it does not suffer from catastrophic forgetting as all the modalities are learned in a single step. 
The fine-tuning approach exhibits positive forward transfer, surpassing even the single-task models on the final task. 
But this comes at the cost of significantly overwriting previous task information. 
It is also more sensitive to the task sequence, when the final modality learned is either RGB or grayscale, then it benefits the other modality learned in the previous steps, as they share a higher degree of similarity.
This trend can also be observed in EWC \cite{EWC}. However, when highly dissimilar modalities are learned in the first and last step, the model fails to learn effectively, as it hinders and prevents overwriting of previous weights.
ILT \cite{ILT} uses knowledge distillation at both the feature and output levels. 
However, aligning features from different modalities can be detrimental, especially for task sequences where the initial and current modalities are vastly different. 
Both RMN \cite{RMN} and our proposed DRMN are more robust and mostly unaffected by the order in the task sequences, achieving consistent results.
Notably, across all three sequences, DRMN outperforms the baseline on the final task.

\subsubsection{InfraParis} \label{sec:experiments:benchmark:infra}

Using the four modalities from the InfraParis \cite{infraParis} dataset, we perform our experiments on the following sequence of tasks:  (\textit{\textbf{IR}} $\rightarrow$ \textit{\textbf{RGB}} $\rightarrow$ \textit{\textbf{Depth}} $\rightarrow$ \textit{\textbf{Gray}}).
Similar to the results from Freiburg Thermal \cite{freiburgThermal} dataset, the single task models form the upper bound.
The addition of another modality in the form of depth images underscores the challenges faced by a single model in handling diverse inputs, as observed in the joint training baseline.
Task order influences fine-tuning results, with better results achieved when the last learned modality (such as gray) complements the previously learned modalities (RGB).
However, due to the separation of task-specific weights, this positive transfer is lower in our DRMN.
Regularization methods such as EWC \cite{EWC} and ILT \cite{ILT} highlight the difficulty of balancing between learning new tasks while retaining previous information.
Both RMN \cite{RMN} and our proposed DRMN achieve comparable results, surpassing the joint training baseline and nearly matching the performance of the single-task models.

\subsection{Shared Weights in RMN for MIL} \label{sec:experiments:shared_weights}
Relevance Mapping Networks (RMNs) preserve knowledge from prior tasks by freezing weights crucial for those tasks, which are identified using the relevance maps.
However, this does not prevent the network from reusing the same weights for new tasks.
It merely forces the network to reuse the old weight values with the newly learned relevance map, preventing any updates to the network's original weights.
We hypothesize that in the context of MIL, where the input modalities are significantly different, sharing weights between tasks can be detrimental and potentially hinder learning. 
To validate this, we explored mechanisms to reduce weight overlap between tasks and force the network to learn independent weights.
The results on Freiburg Thermal \cite{freiburgThermal} are presented in \cref{tab:add_exp:shared_weights}.
Initially, we tried to enforce this constraint through an additional loss term that calculates the overlap between the current task and previous tasks and penalizes it.
The addition of overlap loss to RMN (ORMN) marginally reduces the overlap, which will be discussed further under \cref{sec:experiments:util}, and achieves results similar to the baseline.
This necessitates a more explicit approach to force the network to learn new weights, and we achieve this by masking the previous relevant weights during learning.
We experimented with two variants of this approach: Partial masking with RMN (PRMN), which allows weight sharing in the decoder while preventing overlap in the encoder; and the proposed Disjoint RMN (DRMN), which completely prevents the weights from being shared.
Both methods achieve better results compared to the baseline RMN. 
Notably, DRMN, with stricter weight separation, outperforms the more relaxed PRMN. 
This reinforces our hypothesis that weight sharing can be inhibiting in MIL, leading to conflicts when learning distinct modalities.

\begin{table}[t]
	\caption{Results on the InfraParis \cite{infraParis} dataset after learning all tasks.}
	\begin{adjustbox}{width=\columnwidth}
		\begin{tabular}{c||c|c|c|c|c}
			\boldhline
			\multirow{2}{*}{\textbf{Method}} &\multicolumn{5}{c}{\textbf{IR→RGB→Depth→Gray}} \\ \cline{2-6}
			& IR & RGB & Depth & Gray & \textbf{Avg}  \\ \boldhline
			
			\textbf{Single Task} 	& 38.11 & 62.01  & 55.25 & 61.59 & \textit{54.24}      \\ \hline
			\textbf{Joint Training} & 31.86 & 61.65  & 34.64 & 61.18 & \textit{47.33}      \\ \boldhline
			\textbf{Fine Tuning} 	& 24.84 & 60.67  & 16.19 & 58.83 & \textit{40.13}      \\ \hline
			\textbf{EWC} \cite{EWC}	& 26.72 & 52.44  & 26.90 & 52.37 & \textit{39.61}      \\ \hline
			\textbf{ILT} \cite{ILT}	& 35.79 & 28.38  & 24.62 & 27.85 & \textit{29.16}      \\ \hline
			\textbf{RMN} \cite{RMN}	& 39.85 & 55.21  & 50.10 & 50.14 & \textit{48.82}      \\ \hline
			\textbf{DRMN} (Ours) 	& 39.03 & 53.81  & 50.11 & 52.76 & \textit{\textbf{48.92}}      \\ \boldhline
			
		\end{tabular}
	\end{adjustbox}
	\label{tab:infrapris}	
\end{table}

\begin{table*}[t]
	\centering
	\caption{Results highlighting the influence of weight sharing for three MIL tasks on Freiburg Thermal \cite{freiburgThermal} dataset. From top to bottom, the methods increase in how much separation of neural connections across tasks is enforced (\cf \cref{sec:experiments:shared_weights}).}
	\begin{tabular}{c||c|c|c|c||c|c|c|c||c|c|c|c}
		\boldhline
		\multirow{2}{*}{\textbf{Method}} & \multicolumn{4}{c||}{\textbf{RGB→IR→Gray}}  & \multicolumn{4}{c||}{\textbf{Gray→RGB→IR}} & \multicolumn{4}{c}{\textbf{IR→Gray→RGB}} \\ \cline{2-13}
		& RGB & IR & Gray & \textbf{Avg} & Gray & RGB & IR & \textbf{Avg} & IR & Gray & RGB & \textbf{Avg} \\ \boldhline
		\textbf{RMN} \cite{RMN}  & 73.13  & 55.01  & 68.29  & \textit{65.48}  & 71.09  & 72.82  & 53.57  & \textit{65.83} & 55.10  & 68.90  & 69.46 &  \textit{64.49}    \\ \hline
		\textbf{ORMN} & 73.07  & 54.75  & 67.66  & \textit{65.16}  & 71.06  & 72.53  & 52.79  & \textit{65.46} & 55.11  & 69.02  & 69.00 &  \textit{64.38}       \\\hline
		\textbf{PRMN} & 73.18  & 54.94  & 69.13  & \textit{65.75}  & 71.14  & 72.46  & 53.46  & \textit{65.69} & 55.00  & 70.51  & 70.77 &  \textit{65.43}       \\ \hline
		\textbf{DRMN}  & 73.21  & 54.95  & 69.38  & \textit{\textbf{65.85}}  & 71.12  & 72.61  & 54.12  & \textit{\textbf{65.95}} & 54.97  & 70.56  & 71.19 &  \textit{\textbf{65.57}} \\ \boldhline
	\end{tabular}
	\label{tab:add_exp:shared_weights}	
\end{table*}

\subsection{Task Overlap and Network Utilization} \label{sec:experiments:util}

\begin{table*}[t]
	\caption{Analysis of task overlap and network utilization for task sequence (\textit{\textbf{IR}} $\rightarrow$ \textit{\textbf{Gray}} $\rightarrow$ \textit{\textbf{RGB}}) using Freiburg Thermal \cite{freiburgThermal} dataset.}
	\begin{adjustbox}{width=\textwidth}
		\begin{tabular}{c||ccc||cccccc||cc}
			\boldhline
			\multirow{3}{*}{\textbf{Method}} & \multicolumn{3}{c||}{\textbf{Task-wise Utilization}} & \multicolumn{6}{c||}{\textbf{Pairwise Overlap and Utilization}} & \multicolumn{2}{c}{\textbf{Overall}} \\ \cline{2-12} 
			& \multicolumn{1}{c|}{\multirow{2}{*}{IR}} & \multicolumn{1}{c|}{\multirow{2}{*}{Gray}} & \multirow{2}{*}{RGB} & \multicolumn{2}{c|}{IR \& Gray} & \multicolumn{2}{c|}{IR \& RGB} & \multicolumn{2}{c||}{Gray \& RGB} & \multicolumn{1}{c|}{\multirow{2}{*}{\begin{tabular}[c]{@{}c@{}}Overlap \\ Weights\end{tabular}}} & \multirow{2}{*}{\begin{tabular}[c]{@{}c@{}}Network \\ Utilization\end{tabular}} \\ \cline{5-10}
			& \multicolumn{1}{c|}{} & \multicolumn{1}{c|}{} &  & \multicolumn{1}{c|}{Overlap} & \multicolumn{1}{c|}{Util} & \multicolumn{1}{c|}{Overlap} & \multicolumn{1}{c|}{Util} & \multicolumn{1}{c|}{Overlap} & Util & \multicolumn{1}{c|}{} &  \\ \boldhline
			
			\textbf{RMN} \cite{RMN} & \multicolumn{1}{c|}{47.79} & \multicolumn{1}{c|}{47.78} & 47.80 & \multicolumn{1}{c|}{22.85} & \multicolumn{1}{c|}{72.73} & \multicolumn{1}{c|}{22.84} & \multicolumn{1}{c|}{72.76} & \multicolumn{1}{c|}{22.85} & 72.73 & \multicolumn{1}{c|}{46.68} & 85.77 \\ \hline
			
			\textbf{ORMN} & \multicolumn{1}{c|}{47.79} & \multicolumn{1}{c|}{47.68} & 47.66 & \multicolumn{1}{c|}{22.78} & \multicolumn{1}{c|}{72.69} & \multicolumn{1}{c|}{22.77} & \multicolumn{1}{c|}{72.68} & \multicolumn{1}{c|}{22.72} & 72.62 & \multicolumn{1}{c|}{46.55} & 85.72 \\ \hline
			
			\textbf{PRMN} & \multicolumn{1}{c|}{47.79} & \multicolumn{1}{c|}{24.95} & 16.11 & \multicolumn{1}{c|}{\phantom{0}0.00} & \multicolumn{1}{c|}{72.74} & \multicolumn{1}{c|}{\phantom{0}2.02} & \multicolumn{1}{c|}{61.88} & \multicolumn{1}{c|}{\phantom{0}1.06} & 40.00 & \multicolumn{1}{c|}{\phantom{0}3.08} & 85.77 \\ \hline
			
			\textbf{DRMN} & \multicolumn{1}{c|}{47.79} & \multicolumn{1}{c|}{24.95} & 13.03 & \multicolumn{1}{c|}{\phantom{0}0.00} & \multicolumn{1}{c|}{72.74} & \multicolumn{1}{c|}{\phantom{0}0.00} & \multicolumn{1}{c|}{60.83} & \multicolumn{1}{c|}{\phantom{0}0.00} & 37.99 & \multicolumn{1}{c|}{\phantom{0}0.00} & 85.78 \\ \boldhline
		\end{tabular}
		\label{tab:add_exp:util}
	\end{adjustbox}
\end{table*}

We previously highlighted the importance of learning independent and task-specific weights in \cref{sec:experiments:shared_weights} for MIL.
In this section, we examine how different methods utilize the network and share weights across tasks of Freiburg Thermal \cite{freiburgThermal}, highlighting their influence on the results presented in \cref{tab:add_exp:shared_weights}.
From \cref{tab:add_exp:util}, we can observe that RMN consistently has higher network utilization for each task, with nearly one-fourth of the weights shared between any two tasks and nearly half the weights shared across all tasks.
ORMN, which uses overlap loss to deter weight sharing, exhibits a slight decrease in the percentage of shared weights. 
However, this marginal reduction has negligible impact on the performance of the incrementally learned modalities.
PRMN, which utilizes partial masking, demonstrates a significant decrease in shared weights, leading to improved learning on new modalities.    
Using independent weights raises concerns about limited network capacity for future tasks. 
This concern is heightened with our disjoint RMN (DRMN) approach, which completely prevents weight sharing.
However, \cref{tab:add_exp:util} shows DRMN efficiently learns new tasks with similar network utilization despite having fewer available weights, alleviating concerns about network capacity exhaustion. 
An analog experiment on InfraParis \cite{infraParis} indicates the same trend.
The exact utilization on that dataset is detailed in the supplementary material.

\subsection{Efficiency and Scalability} \label{sec:experiments:eff}
Compared to growing architectures, DRMN maintains a constant size for any number of tasks.
Duplicating encoders, decoders, or entire networks in each task introduces a significant linear growth of the overall model. 
Similarly, the original RMNs require one full relevance map for all weights for each task.
However, the disjoint property of DRMNs allows for an efficient implementation that stores the relevance maps for all tasks in a single data structure.

The overhead of selecting and loading the appropriate relevance maps is also constant and negligibly small compared to the computation within the network.

\section{Conclusion} \label{sec:conclusion}
This work introduces a new paradigm called modality-incremental learning (MIL).
In contrast to existing incremental learning settings where the input distribution comes from the same visual modality, MIL addresses a larger domain gap between tasks, as the modalities can vary significantly. 
Consequently, existing continual learning methods and baselines fall short in handling multiple modalities, necessitating the development of tailored and dynamic approaches for MIL. 
We build upon the Relevance Mapping Network (RMN).
Unlike dynamically growing architecture methods that raise scalability concerns, RMNs use a fixed network architecture and relevance maps to incrementally adapt to new tasks. 
We introduce a crucial modification with our Disjoint RMN (DRMN) by strictly separating neural connections across tasks. 
This approach demonstrates improved learning across modalities by reducing conflicts between them, though keeping the overall network utilization at a comparable level.
For future work, we plan to implement an adaptive regularization term for the overlap between task-specific relevance maps that considers the similarity between modalities.

\section*{Acknowledgments}
This work was partially funded by the Federal Ministry of Education and Research Germany under the projects DECODE (01IW21001) and COPPER (01IW24009) and partially under the EU project ExtremeXP (GA Nr 101093164).

{\small
\bibliographystyle{ieee_fullname}
\bibliography{bib}
}

\newpage

\twocolumn[
    \centering
    \Large
    \bf
    Supplementary Material
    \vskip 1em
]

\renewcommand\thesection{\Alph{section}}
\setcounter{section}{0}

\section{Overview}
In this supplementary material to our paper \textit{Modality-Incremental Learning with Disjoint Relevance Mapping Networks for Image-based Semantic Segmentation}, we show the impact of forgetting on previously learned modalities, test the robustness of Disjoint Relevance Mapping Networks (DRMNs) against variation of the pruning parameter $\mu$, and list the exact utilization of network connections for the experiment on InfraParis \cite{infraParis}.

\section{Task-wise Evaluation}
To quantify the amount of forgetting due to the incremental learning of different modalities, \cref{supp:tab:taskwise} provides the mIoU for each modality through the learning sequence.
\Ie, each known modality is evaluated after each task.
This way, the mutual negative influence of the modalities can be measured.
With regularization-based approaches such as EWC \cite{EWC} and ILT \cite{ILT}, the model learns optimally during the initial step as expected. 
However, when learning other modalities incrementally, EWC prevents overwriting important parameters from the previous modalities, hindering its learning on the new modality. 
On the other hand, ILT which uses distillation, exhibits better performance on the initial task compared to EWC. 
However, the performance on new modalities is significantly worse due to the diverse nature of the modalities.
In RMN \cite{RMN} and the proposed DRMN, even for the initial modality the results are slightly lower compared to the single-task models.
This is due to the use of relevance maps, which preserve network capacity for future tasks by not utilizing the entire network capacity at each step.
This approach effectively preserves information and completely mitigates catastrophic forgetting, ensuring that performance on previously learned modalities remains consistent over the sequence of tasks.
Additionally, with DRMN, isolating parameters and learning task-specific weights enhances the learning of new modalities, as evident in improved performance on both Gray and RGB tasks.

\begin{table*}[t]
    \centering
    \caption{Results on Freiburg Thermal \cite{freiburgThermal} for the original RMN \cite{RMN} and our proposed DRMN after each step of training the sequence (\textit{\textbf{IR}} $\rightarrow$ \textit{\textbf{Gray}} $\rightarrow$ \textit{\textbf{RGB}}).}
    \label{supp:tab:taskwise}
        \begin{tabular}{c||c||c|c||c|c|c}
            \boldhline
            \multirow{2}{*}{Method} & $M_0$ (IR) & \multicolumn{2}{c||}{$M_1$ (Gray)} & \multicolumn{3}{c}{$M_2$ (RGB)} \\
            & IR & IR & Gray & IR & Gray & RGB \\
            \boldhline
            \textbf{Fine Tuning} & 59.56 & 07.44 & 74.10 & 07.12 & 73.17 & 75.24 \\
            \boldhline 
            \textbf{EWC} \cite{EWC} & 59.75 & 06.89 & 58.04 & 10.56 & 59.47 & 61.69 \\ \hline
            \textbf{ILT} \cite{ILT} & 59.56 & 20.39 & 21.08 & 20.68 & 22.57 & 23.53 \\ \hline
            \textbf{RMN} \cite{RMN} & 55.30 & 55.18 & 68.85 & 55.10 & 68.90 & 69.46 \\ \hline
            \textbf{DRMN} (Ours)    & 55.30 & 55.16 & 70.61 & 54.97 & 70.56 & 71.19 \\
            \boldhline
    \end{tabular}

\end{table*}

\begin{table*}[t]
	\centering
	\caption{Results and task-wise network utilization on Freiburg Thermal \cite{freiburgThermal} for the original RMN \cite{RMN} and our proposed DRMN with varying pruning parameters.}
	\label{supp:tab:mu}
	\begin{tabular}{c||c||c|c|c|c||c|c|c||c||c}
		\boldhline
		\multirow{2}{*}{\textbf{Method}} & \multirow{2}{*}{\textbf{Prune}} & \multicolumn{4}{c||}{\textbf{Results (mIoU)}} & \multicolumn{3}{c||}{\textbf{Task Utilization (\%)}} &  \multicolumn{2}{c}{\textbf{Overall (\%)}} \\ \cline{3-11}
		& $\mu$ & IR & Gray & RGB & \textbf{Avg} & IR & Gray & RGB & {\begin{tabular}[c]{@{}c@{}}Shared \\ Weights\end{tabular}} & {\begin{tabular}[c]{@{}c@{}}Network \\ Utilization\end{tabular}}\\ \boldhline 
		\multirow{3}{*}{\textbf{RMN} \cite{RMN}} &
		0.5 & 55.02 & 69.06 & 68.96 & 64.35 & 49.91 & 49.85 & 49.83 & 49.80 & 87.39 \\ \cline{2-11}
		& 0.6 & 55.10 & 68.90 & 69.46 & 64.49 & 47.79 & 47.78 & 47.80 & 46.68 & 85.77 \\ \cline{2-11}
		& 0.7 & 54.25 & 68.33 & 68.66 & 63.75 & 49.10 & 49.03 & 49.08 & 48.62 & 86.78 \\ \boldhline
		\multirow{3}{*}{\textbf{DRMN} (Ours)} &
		0.5 & 55.23 & 70.64 & 71.21 & 65.69 & 49.91 & 24.94 & 12.50 & \phantom{0}0.00 & 87.35 \\ \cline{2-11}
		& 0.6 & 54.97 & 70.56 & 71.19 & 65.57 & 47.79 & 24.95 & 13.03 & \phantom{0}0.00 & 85.78 \\ \cline{2-11}
		& 0.7 & 54.34 & 70.69 & 71.05 & 65.36 & 49.10 & 24.93 & 12.73 & \phantom{0}0.00 & 86.76 \\ \boldhline
	\end{tabular}
\end{table*}

\begin{table*}[!htbp]
    \centering
    \caption{Network utilization on InfraParis \cite{infraParis} for the original RMN \cite{RMN} and our proposed DRMN on the task sequence (\textit{\textbf{IR}} $\rightarrow$ \textit{\textbf{RGB}} $\rightarrow$ \textit{\textbf{Depth}} $\rightarrow$ \textit{\textbf{Gray}}).}
    \label{supp:tab:util}
    \begin{tabular}{c||c|c|c|c||c||c}
			\boldhline
			\multirow{3}{*}{\textbf{Method}} &\multicolumn{4}{c||}{\textbf{Task Utilization (\%)}} &  \multicolumn{2}{c}{\textbf{Overall (\%)}} \\ \cline{2-7}
			& IR & RGB & Depth & Gray  & {\begin{tabular}[c]{@{}c@{}}Shared \\ Weights\end{tabular}} & {\begin{tabular}[c]{@{}c@{}}Network \\ Utilization\end{tabular}}\\ \boldhline
			
			\textbf{RMN} \cite{RMN} & 49.52 & 49.54 & 49.49 & 49.54 & 68.02 & 93.50 \\ \hline
			\textbf{ORMN}           & 49.52 & 49.46 & 49.40 & 49.41 & 67.91 & 93.47 \\ \hline
			\textbf{PRMN}           & 49.52 & 27.16 & 15.84 & 10.16 &  5.97 & 93.48 \\ \hline
			\textbf{DRMN} (Ours)    & 49.52 & 24.98 & 12.61 &  6.37 &  0.00 & 93.49 \\ \boldhline
		\end{tabular}
\end{table*}

\section{Relevance Map Pruning}
To recall, the hyperparameter $\mu$ defines the threshold at which network weights (connections) are considered relevant.
Any connection below this threshold will be pruned after every epoch above 50.
The values in the relevance map of the pruned connections will be permanently set to zero for this task, removing the influence of that connection entirely.
The unused connections might be used in a later task, though.
The network's weights for relevant connections will be frozen.
The default value for $\mu$ is $0.6$. 
In \cref{supp:tab:mu}, we show the results for a threshold of $0.5$ and $0.7$.
The variation of $\mu$ in both directions indicates a high robustness of DRMN in this regard.
Same holds for the original RMN.
Interestingly, we point out that varying the pruning parameter has no significant impact on the sparsity (utilization) of neural connections.

\section{Task Utilization on InfraParis}
One of our claims in the main paper is that despite the strict separation of task-specific connections, the network's capacity is not exceeded faster than with regular RMNs.
To back this claim further, we have also computed the network utlization for the four tasks of InfraParis \cite{infraParis}.
The result is shown in \cref{supp:tab:util}.
For a description of ORMN and PRMN, we refer to Sec.~5.5 of the main paper.
It is striking that even with just about 6~\% of the network's overall connections, the final task can be learned even better than with RMN, which uses about half of all weights.
Another remarkable observation is that each task approximately consumes half of the remaining connections in DRMN.

\end{document}